\documentclass{article}

\PassOptionsToPackage{numbers, compress}{natbib}

\usepackage[preprint]{neurips_2026}

\usepackage[utf8]{inputenc} 
\usepackage[T1]{fontenc}    
\usepackage{hyperref}       
\usepackage{url}            
\usepackage{booktabs}       
\usepackage{multirow}       
\usepackage{amsfonts}       
\usepackage{nicefrac}       
\usepackage{microtype}      
\usepackage{xcolor}         

\usepackage{amsmath}
\usepackage{graphicx}

\newcommand{\method}{CommunityKV}
\newcommand{\argtopk}{\operatorname*{arg\,top\text{-}\kappa}}
\newcommand{\topk}{\operatorname*{top\text{-}\kappa}}

\title{CommunityKV: Efficient Long-Context Decoding via Graph Partitioning}

\author{Joe McKenna \\
  Amazon AGI \\
  \texttt{joemcke@amazon.com} \\
  \And
  Anastasios Alexandridis \\
  Amazon AGI \\
  \texttt{aanst@amazon.com} \\
  \And
  Nathan Susanj \\
  Amazon AGI \\
  \texttt{nsusanj@amazon.com} \\
  \And
  Jing Liu \\
  Amazon AGI \\
  \texttt{jlmk@amazon.com} \\
}

\begin{document}

\maketitle

\begin{abstract}
Scaling Transformers to long contexts is constrained by the quadratic cost of self-attention and the linear growth of key-value cache memory transfer. Sparse attention mitigates this by retrieving only relevant tokens, but current approaches either require large-scale training or, within the training-free regime, rely on semantically coarse heuristics or expensive clustering that is difficult to update efficiently during decoding. We introduce CommunityKV, a framework that formulates sparse attention as a community detection problem. CommunityKV constructs a token graph from the $QK^T$ scores already computed during standard prefill, and partitions the graph into communities to enable retrieval of semantically coherent token groups. A local update rule assigns newly generated tokens to communities in constant time, enabling sparse retrieval throughout streaming decoding without global re-partitioning. We evaluate CommunityKV on Qwen3 and Llama-3.1 models across three long-context benchmarks. With one graph per query head, CommunityKV delivers up to $1.25\times$ the end-to-end generation throughput of dense attention, while query-group graph aggregation yields up to $1.71\times$ with comparable accuracy.
\end{abstract}

\section{Introduction}

The capacity for Transformers~\cite{vaswani2017attention} to process long context inputs powers a new generation of applications, ranging from repository-level code understanding~\citep{bairi2023codeplan, rando2025longcodebench} and long-horizon agentic planning~\cite{erdogan2025plan} to multimodal reasoning~\cite{wang2025multimodal}. However, as the context length grows, standard self-attention encounters prohibitive bottlenecks: the quadratic cost of attention and the linear growth of key-value (KV) cache memory transfer.

Despite the increasing KV cache size for modern applications, empirical studies demonstrate that token prediction accuracy is driven by only a small fraction of relevant past tokens~\cite{zhang2023h2o, ge2023model}. Sparse attention methods attempt to exploit this redundancy by loading only a portion of the KV cache at each generation step~\cite{tay2022efficient, nawrot2025sparse}. Recent architectural methods such as DeepSeek Sparse Attention~\cite{deepseekv32} integrate sparsity into the training objective, but require large-scale training. Within the training-free regime, current approaches force a compromise: they either employ retrieval heuristics that suffer from unacceptable accuracy degradation~\cite{tang2024quest} or rely on heavy preprocessing of the KV cache that is difficult to update during decoding~\cite{hooper2025multipole}.

In this work, we propose \method, a framework that formulates training-free sparse attention as community detection on a graph induced from attention scores. \method~leverages the structure already present in the $QK^\top$ matrix to identify semantically coherent token groups. A local update rule extends the partition to newly generated tokens in constant time, enabling streaming decoding without re-clustering.\footnote{Code is available at \url{https://github.com/amazon-science/community-kv}.}

Our contributions are as follows:
\begin{itemize}
\item \textbf{Formulation:} We formulate sparse attention as community detection on a token graph induced from $QK^\top$ scores. A combined graph captures both direct query-key dependencies and second-order co-attention structure, enabling retrieval of semantically coherent token groups without auxiliary embeddings or vector-space clustering. A local update rule extends the partition to newly generated tokens, supporting streaming decoding without global re-partitioning.
\item \textbf{Complexity analysis:} We show that graph construction is amortized within the standard prefill matrix multiplication, and partitioning adds only $\mathcal{O}(n \log n + nd)$. During decoding, community retrieval is sub-linear at $\mathcal{O}(\sqrt{n}\,d)$ per step, and adding a new token to the partition is $\mathcal{O}(d)$, independent of sequence length, achieving sub-linear retrieval with constant-time updates simultaneously.
\item \textbf{Empirical results:} We evaluate CommunityKV across two model families and three long-context benchmarks. In end-to-end measurements, the per-query-head configuration achieves $1.08\text{--}1.25\times$ throughput with 17--29\% additional peak memory, while query-group aggregation achieves $1.39\text{--}1.71\times$ throughput with 0.5--8.5\% additional peak memory and comparable accuracy.
\end{itemize}

\section{Related work}

\paragraph{Architectural vs.\ training-free sparse attention}

Recent methods integrate sparsity directly into the training objective, achieving large throughput gains. DeepSeek Sparse Attention (DSA)~\cite{deepseekv32} scores queries against lightweight indexer keys to select a top-$k$ subset of past tokens for full attention, while Native Sparse Attention (NSA)~\cite{yuan2025native} combines block-level token compression with fine-grained token selection. However, these approaches couple sparsity to model training: the upfront cost amortizes only at very high inference volumes, and the resulting sparsity pattern is fixed at training time, limiting retrieval quality to the training distribution.

CommunityKV targets the complementary setting: training-free sparse attention. Within the training-free regime, existing approaches reduce KV cache traffic via two strategies. KV cache compression methods~\cite{zhang2023h2o, li2024snapkv, liu2023scissorhands} enforce strict memory ceilings through token eviction or merging, but are inherently lossy: once discarded, information cannot be recovered. Selective retrieval methods instead preserve the full KV cache in memory and dynamically load only relevant tokens at each decoding step. The effectiveness of retrieval-based methods depends on the design of the sparsity mask, which typically employs either fixed or dynamic patterns.

\paragraph{Fixed-pattern sparse attention}

Early work on sparse attention relied on predefined, query-independent sparsity patterns. The Sparse Transformer~\cite{child2019generating} introduced factorized attention with strided patterns, achieving sub-quadratic $\mathcal{O}(n\sqrt{n}\,d)$ complexity. Longformer~\cite{beltagy2020longformer} and BigBird~\cite{zaheer2020big} combined sliding windows, dilated patterns, and global tokens to achieve linear $\mathcal{O}(nd)$ complexity. LongNet~\cite{ding2023longnet} extended dilated attention to billion-token sequences, and StreamingLLM~\cite{xiao2024efficientstreaminglanguagemodels} showed that retaining initial ``attention sinks'' stabilizes sliding window attention. Because these patterns are fixed in advance, they miss relevant tokens that fall outside the predefined structure.

\paragraph{Dynamic sparse attention}

Dynamic sparse attention frames the problem as query-dependent retrieval, typically using auxiliary data structures to index the KV cache and select relevant subsets at each step. A first family uses heuristics over contiguous token blocks. Quest~\cite{tang2024quest} tracks channel-wise min-max values to estimate query-key overlap, while a similar training-based method MoBA~\cite{lu2025moba} applies block-wise gating to filter context chunks. These methods avoid heavy auxiliary data structures but treat blocks of contiguous tokens as the retrieval unit: a high-scoring outlier within a block triggers retrieval of the entire block, including unrelated neighbors. A second family uses Locality Sensitive Hashing~\cite{kitaev2020reformer, zandieh2023kdeformer, chen2024magicpig, han2023hyperattention} to bucket tokens using random projections. Because these projections are data-independent, they fail to account for the distribution of key embeddings; achieving high recall requires scaling the number of hash tables, which re-introduces significant overhead. A third family applies $k$-means clustering to the key or value vectors~\cite{roy2021efficient, hooper2025multipole, vyas2020fast, wang2021cluster}. These methods produce semantically coherent clusters, but $k$ must be chosen a priori and centroids drift as new tokens are generated, forcing a trade-off between stale clusters and expensive re-clustering.

\paragraph{Comparison to GraphKV}

GraphKV~\cite{li2025graphkv} is the most closely related work, constructing a sparse attention graph and applying a ``decay signal propagation'' algorithm to evict semantically redundant tokens. The result is a compressed cache that every subsequent decoding step queries against. CommunityKV adopts the opposite philosophy. Instead of treating semantic similarity as a signal to be decayed, we treat it as the basis for community formation: we retain the full KV cache and dynamically retrieve communities for each query, allowing the model to access locally dense semantic structure without permanently deleting context.

\section{Method}

\begin{figure}
\centering
\includegraphics[width=\linewidth]{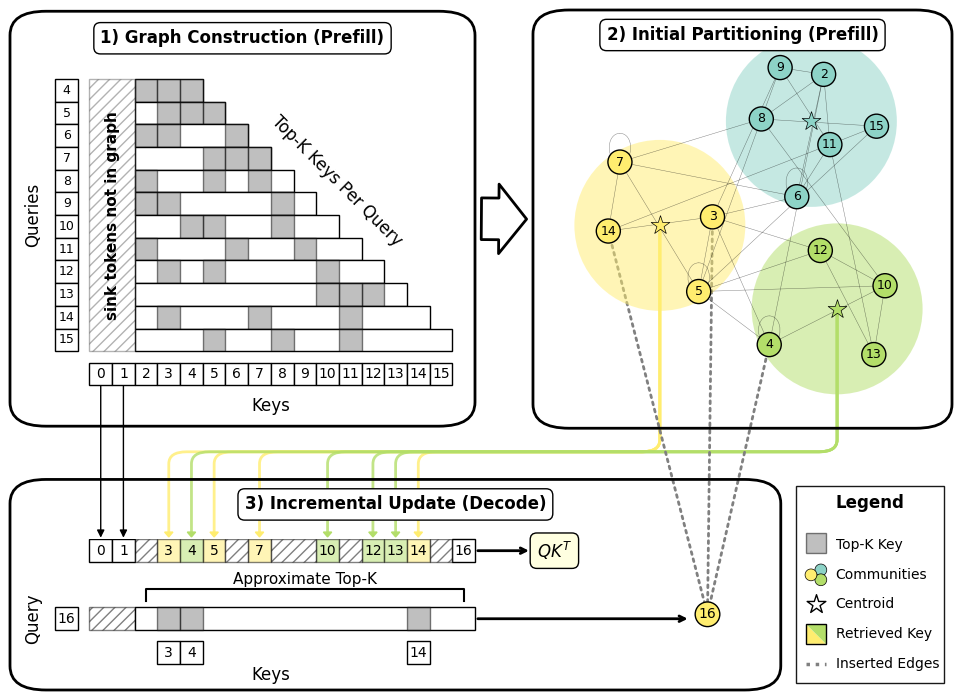}
\caption{Overview of the \method~framework. \textbf{(1) Graph Construction.} During prefill, we induce a sparse graph from the attention matrix by connecting each query to its top-$\kappa$ keys (gray cells, $\kappa=3$ shown). Sink tokens (hatched) are excluded from the top-$\kappa$ pool. \textbf{(2) Initial Partitioning.} We partition the graph into semantic communities (colored regions) using the Leiden algorithm and compute a centroid (star) per community. \textbf{(3) Incremental Update.} At each decoding step (token 16 shown), we score the query against community centroids to retrieve a candidate subset, run exact attention on this subset together with sink tokens, and use the resulting top-$\kappa$ keys to update the new token's edges and assign it to a community in constant time. For clarity, only direct attention $w^{(1)}$ relations are depicted; this corresponds to $\lambda=1$ in the combined graph.}
\label{fig:method}
\end{figure}

Our approach has three phases (Figure~\ref{fig:method}): (1) graph construction, where we build a sparse token graph from attention scores during prefill; (2) initial partitioning, where we cluster the graph into semantic communities; and (3) incremental update, where newly generated tokens are assigned to communities in constant time during decoding.

\subsection{Graph construction}
\label{sec:graph_construction}

We construct a graph where nodes represent tokens and edges encode two kinds of relations: direct attention, where one token attends strongly to another, and co-attention, where two tokens are jointly attended by a common query. Let $\alpha \in [0,1]^{n \times n}$ be the attention score matrix over an $n$-token context, where $\alpha_{ij}$ is the score of query $i$ attending to key $j$, and let $\mathcal{K}_i$ be the $\topk$ key positions for query $i$:
\begin{equation}
\label{eq:topkeyset}
\mathcal{K}_i = \argtopk_{j =|\mathcal{S}|+1,\ldots,i} \alpha_{ij}.
\end{equation}
We exclude the set of initial ``sink token'' positions $\mathcal{S}$, as they exhibit disproportionately high attention scores regardless of semantic relevance~\cite{xiao2023efficient}. 

\paragraph{Direct attention graph}

The direct attention graph captures explicit query-key dependencies: a directed edge $i \to j$ records that query $i$ attends strongly to key $j$:
\begin{equation}
\label{eq:direct_attn}
w^{(1)}_{ij} = \alpha_{ij}\, \mathbf{1}\left(j \in \mathcal{K}_i\right)
\end{equation}
where $\mathbf{1}(\cdot)$ is the indicator function.

\paragraph{Co-attention graph}

If two keys are strongly attended by the same query, they play similar roles in that query's context. The co-attention graph captures this by linking keys that co-occur in the top-$\kappa$ set of a common query:
\begin{equation}
\label{eq:co_attn}
w^{(2)}_{ij} = \sum_{m=|\mathcal{S}|+1}^n \alpha_{mi} \, \alpha_{mj} \, \mathbf{1}\left(\{i,j\} \subseteq \mathcal{K}_m\right).
\end{equation}

\paragraph{Combined graph}

We combine the two adjacency matrices into a single symmetric adjacency matrix:
\begin{equation}
w = \frac{\lambda}{2}\left[w^{(1)} + (w^{(1)})^{\top}\right] + (1-\lambda)\, w^{(2)},
\end{equation}
with $0\le\lambda\le1$ controlling the balance between direct and co-attention relations. By construction, the graph is sparse: each of the $n$ queries contributes $\kappa$ edges to $w^{(1)}$ and at most $\kappa^2$ edges to $w^{(2)}$, so the total edge count satisfies $E=\mathcal{O}(n\kappa(1+\kappa)) = \mathcal{O}(n)$ for constant $\kappa$.

\paragraph{Theoretical justification}

The combined graph $w$ admits a dynamical interpretation that motivates graph partitioning, grounded in the mathematical structure of self-attention. Because $\alpha$ is row-stochastic, it defines the one-step transition kernel of a discrete Markov chain over the token sequence. Symmetrizing as $(\alpha + \alpha^\top)/2$ and combining with the second-order affinity matrix $\alpha^\top \alpha$ yields a symmetric kernel of a time-reversible random walk, and sparsifying via top-$\kappa$ selection produces $w$, preserving the dominant transition paths with bounded sparsity ($E = \mathcal{O}(n)$).

In this framework, token communities correspond to metastable states of the walk: groups of tokens the walk is unlikely to escape over time. We partition the graph into these communities by maximizing modularity~\cite{newman2006modularity}, which on the symmetric graph defined by $w$ is equivalent to maximizing the linearized Markov Stability of the induced walk~\cite{lambiotte2014random}. In practice, we employ the Leiden algorithm~\cite{traag2019louvain} to perform this graph partitioning.

\subsection{Initial partitioning}

\paragraph{Leiden community detection}

The Leiden algorithm~\cite{traag2019louvain} iterates three phases: fast local moving, refinement, and aggregation, to maximize modularity:
\begin{equation}
\label{eq:modularity_score}
Q=\frac1{2m}\sum_{ij}\left(w_{ij} - \gamma\frac{w_iw_j}{2m}\right)\mathbf{1}(c_i=c_j),
\end{equation}
where $m = \frac{1}{2}\sum_{ij} w_{ij}$ is the total graph weight, $w_i = \sum_j w_{ij}$ is the weighted degree of token $i$, and $c_i$ is the community label of token $i$. The resolution parameter $\gamma$ controls community granularity: larger values favor smaller, denser communities. Leiden improves upon Louvain~\cite{blondel2008fast} by guaranteeing connected communities.

We choose Leiden over alternative community detection algorithms for several reasons. Leiden empirically exhibits $\mathcal{O}(E)$ runtime~\cite{sahu2024fast, traag2019louvain} and few iterations required for convergence; \cite{blondel2008fast} report fewer than 5 across all tested networks for the analogous Louvain algorithm. We cap iterations at $r = \lfloor \log_{10} n \rfloor$, yielding $r \in [4, 7]$ for our evaluation contexts (10k--10M tokens). With graph sparsity bounded at $E = \mathcal{O}(n)$, graph partitioning costs $\mathcal{O}(rE) = \mathcal{O}(n \log n)$. We then compute community centroids in a single $\mathcal{O}(nd)$ pass, yielding a clustering cost of $\mathcal{O}(n \log n + nd)$ (Table~\ref{tab:complexity}). By contrast, Multipole Attention assigns $n$ tokens to $\mathcal{O}(n)$ centroids via $k$-means, yielding $\mathcal{O}(n^2 d)$ clustering cost. Because pairwise similarity is already captured in the graph edges via $QK^\top$, our clustering operates on scalar weights rather than $d$-dimensional vectors.

Leiden's local-moving phase is embarrassingly parallel; each node's modularity-gain computation $\Delta Q$ (Eq.~\ref{eq:deltaq}) depends only on its neighbors and community-level aggregates, so all nodes can be evaluated simultaneously. Recent parallel implementations exploit this structure to achieve a processing rate of 403M edges/s on a 3.8B-edge graph using 64 CPU threads~\cite{sahu2024fast}, and NVIDIA's cuGraph~\cite{fender2022rapids} reports significant speedup over CPU baselines.

Leiden admits incremental update schemes that avoid global re-partitioning when edges are added~\cite{sahu2024heuristic, lin2026efficient}. Our local update rule (Section~\ref{sec:incremental_update}) is a domain-specific specialization of this idea, exploiting the structure of autoregressive decoding to assign new tokens in $\mathcal{O}(d)$ time per step.

\subsection{Incremental update}
\label{sec:incremental_update}

\paragraph{Community retrieval}

During decoding, we retrieve relevant tokens by scoring the latest query $q_i$ against community centroids and aggregating positions from the top-ranking communities into a candidate set $\mathcal{R}_i$. We then form the sparse context $\mathcal{S} \cup \mathcal{R}_i \cup \{i\}$ by augmenting $\mathcal{R}_i$ with the attention sinks $\mathcal{S}$ and the current query position $i$, truncating to a fixed token budget. Because our graph satisfies $E=\mathcal{O}(n)$, the modularity resolution limit~\cite{fortunato2007resolution} implies the number of communities scales as $\mathcal{O}(\sqrt{n})$ (verified empirically in Figure~\ref{fig:community_scaling}), making retrieval cost $\mathcal{O}(\sqrt{n}\,d)$ (Table~\ref{tab:complexity}).

\paragraph{Edge approximation}

To integrate a new token $i$ into the graph, we approximate its edge weights using only the retrieved set. We compute approximate attention scores and top-$\kappa$ keys over $\mathcal{R}_i \cup \{i\}$:
\begin{equation}
\label{eq:approx_topk}
\begin{split}
\tilde{\alpha}_{ij} &= \text{softmax}(q_i^\top k_j)_{j \in \mathcal{R}_i\cup\{i\}}, \\
\tilde{\mathcal{K}}_i &= \argtopk_{j \in \mathcal{R}_i\cup\{i\}} \tilde{\alpha}_{ij}.
\end{split}
\end{equation}
Substituting these approximations into Eqs.~\eqref{eq:direct_attn} and~\eqref{eq:co_attn} yields the edge updates. We add $\kappa$ direct attention edges $w^{(1)}_{ij} = \tilde{\alpha}_{ij}$ for each $j \in \tilde{\mathcal{K}}_i$, and increment the $\kappa^2$ co-attention edges between pairs $j, k \in \tilde{\mathcal{K}}_i$ by $\Delta w^{(2)}_{jk} = \tilde{\alpha}_{ij}\tilde{\alpha}_{ik}$. 

\paragraph{Greedy community assignment}

After integrating the new token $i$ into the graph, we assign it a community label. We freeze the existing community assignments and consider two options: form a new singleton community $\{i\}$ or join a neighboring community $C \in \{c_j \mid j \in \tilde{\mathcal{K}}_i\}$. We pick the option that maximizes the modularity gain:
\begin{equation}
\label{eq:deltaq}
\Delta Q(i\rightarrow C) \propto \sum_{j\in C\,\cap\,\tilde{\mathcal{K}}_i} w_{ij} - \gamma \frac{w_i w_C}{2m},
\end{equation}
where $w_C = \sum_{j\in C} w_j$ is the total weight of community $C$.

Each decoding step touches at most $\kappa(\kappa+1)$ edges, evaluates $\kappa$ candidate communities, and updates one centroid in $\mathcal{O}(d)$. Since $\kappa$ is a small constant, the total per-step update cost is $\mathcal{O}(d)$, independent of sequence length $n$ (Table~\ref{tab:complexity}).

\begin{table}
\caption{Computational complexity comparison. $n$ is sequence length, $d$ is hidden dimension. Note that our method achieves sub-linear retrieval $\mathcal{O}(\sqrt{n}d)$ without expensive clustering or update costs.}  
\label{tab:complexity}
\centering

\begin{tabular}{lccc}
\toprule
Method  & \textbf{Initial Clustering} & \textbf{Per-step Update} & \textbf{Per-Step Retrieval} \\
\midrule
Quest & $\mathcal{O}(nd)$ & $\mathcal{O}(d)$ & $\mathcal{O}(nd)$ \\
Multipole & $\mathcal{O}(n^2d)$ & $\mathcal{O}(nd)$ & $\mathcal{O}(nd)$ \\
\method & $\mathcal{O}(n\log n + nd)$ & $\mathcal{O}(d)$ & $\mathcal{O}(\sqrt{n}\,d)$ \\
\bottomrule
\end{tabular}
\end{table}

\paragraph{Stability of incremental updates}

Our local update rule maintains partition quality throughout decoding without global re-partitioning. Over a 4096-token generation horizon, modularity remains stable while attention mass decays gracefully, staying above 85\% at a 4096-token budget (Appendix~\ref{app:drift}). The KV cache and the routing graph are decoupled in \method: updates modify only the graph, leaving past key-value states untouched. A single global re-partitioning restores optimal modularity in the graph. Running Leiden every $\mathcal{O}(\sqrt{n})$ steps adds amortized $\mathcal{O}(\sqrt{n} \log n)$ cost per step, making periodic refresh feasible for unbounded generation. Over a shared 16,384-token PG-19~\cite{rae2020compressive} continuation, repartitioning every 512 steps improves CommunityKV's final 512-token window top-1 agreement with dense attention from 68.05\% to 82.81\% (Appendix~\ref{app:fixed_continuations}).

\subsection{Implementation}

We fuse top-$\kappa$ selection into the FlashAttention QK main loop using a WarpSelect-style selector~\cite{johnson2019billion}. For each query row, GPU threads maintain small sorted candidate queues, updated via compare-and-insert as each $b_q \times b_k$ score tile is computed. After the final key tile, the queues are merged with warp shuffles to produce the row's top-$\kappa$ key indices and corresponding weights, which are normalized using the row's final softmax statistics. The total cost is $\mathcal{O}(n^2)$, matching the underlying prefill matrix multiplication (Appendix~\ref{app:fused_topk}). Empirically, this adds 6--8\% to prefill wall time (Appendix~\ref{app:latency_breakdown}).

Leiden partitioning is pipelined with the forward pass: as soon as layer $L$'s attention scores are available, its graphs are dispatched to asynchronous workers while layers $L{+}1, \ldots$ continue executing (Appendix~\ref{app:pipelined_partitioning}). The only exposed latency is the final layer's partition (0.37--0.45s across the Qwen3 family). We provide the fused top-$\kappa$ selection complexity derivation, pipelined graph-partitioning schedule, component-level latency breakdown, and memory accounting in Appendix~\ref{app:impl}.

\section{Experiments}

We evaluate CommunityKV along three axes: (1) accuracy compared to sparse attention baselines on long context benchmarks, (2) end-to-end throughput and peak GPU memory, and (3) ablation studies over key design parameters. Unless otherwise noted, we fix $\kappa=8$, $\lambda=0.5$, exclude the first 10 sink tokens from graph construction, and maintain one independent graph per query head. All models use greedy decoding (single run, deterministic) with a maximum of 128 generated tokens and YaRN RoPE scaling~\cite{peng2023yarn} to extend the model context. To ensure a fair comparison, we tune $\gamma$ to produce a mean community size of 16, matching the baselines' retrieval granularity.

\subsection{Experimental setup}

\subsubsection{Datasets}

\paragraph{LongBench v2} We evaluate on LongBench v2~\cite{bai2025longbench}, a suite of challenging multiple-choice questions spanning code understanding, long-dialogue history, and multi-document QA.
\paragraph{BABILong} We additionally evaluate on BABILong~\cite{kuratov2024babilong} QA1--QA5, a multi-hop reasoning benchmark requiring retrieval of 1--5 supporting facts from long context inputs.
\paragraph{RULER} We further evaluate on RULER~\cite{hsieh2024ruler}, a configurable long-context benchmark spanning retrieval, multi-hop tracing, aggregation, and question answering.

\subsubsection{Models}

\paragraph{Models} We evaluate the Qwen3 family~\cite{yang2025qwen3}, specifically the 4B, 8B, and 14B variants, and Llama-3.1-8B-Instruct~\cite{grattafiori2024llama}.

\subsubsection{Baselines}

\textbf{FlashAttention-2 (Dense)} Exact FlashAttention-2~\cite{dao2023flashattention} serves as the lossless reference for both accuracy and wall-clock latency.

\textbf{GraphKV} As the most closely related graph-based method, GraphKV~\cite{li2025graphkv} constructs a sparse attention graph and permanently evicts low-scoring tokens during prefill. We include it to isolate the benefit of dynamic retrieval over static eviction.

We additionally compare against two training-free retrieval methods that span the efficiency-fidelity spectrum:

\textbf{Quest} Quest~\cite{tang2024quest} partitions consecutive tokens into fixed blocks and tracks channel-wise min-max statistics for retrieval, representing the fast but semantically coarse end of the spectrum.

\textbf{Multipole Attention} Multipole~\cite{hooper2025multipole} clusters keys via $k$-means, representing the high-fidelity end. It serves as a performance target that our lighter approach seeks to match without the extra compute cost.

We additionally compare against direct training-free KV retrieval methods (TokenSelect, FreeKV, and SparQ)~\cite{wu2025tokenselect, liu2025freekv, ribar2024sparq} and eviction/compression methods (H2O, SnapKV, and StreamingLLM) (Appendix~\ref{app:expanded_baselines}).

\subsection{Long context understanding}
\label{sec:results_long_context}

\paragraph{Graph-based retrieval vs.\ KV eviction}
Both \method~and GraphKV construct sparse graphs from attention scores, but differ in how they use the graph: GraphKV evicts tokens permanently, while \method~retrieves dynamically. To isolate this distinction, we compare the two on LongBench v2 with Qwen3-8B, sweeping GraphKV's budget from 4k to 32k tokens (Table~\ref{tab:comparison_graphkv}). GraphKV requires an $8\times$ larger budget (32k tokens) to reach 22.5\% accuracy, while \method~achieves 31.1\% with only 4k tokens, nearly matching the dense baseline. This confirms that semantic structure should guide dynamic retrieval rather than permanent context deletion.

\begin{table}
\caption{Accuracy on LongBench v2 across token budgets (4k--32k) for GraphKV and CommunityKV. Dynamic community retrieval matches dense accuracy at $8\times$ lower budget than static eviction.}
\label{tab:comparison_graphkv}
\centering
\begin{tabular}{lrrrr}
\toprule
\multicolumn{5}{c}{\textit{Qwen3-8B} (Dense: 31.4)} \\
\midrule
\textbf{Method} & 4096 & 8192 & 16384 & 32768 \\
\midrule
GraphKV & 3.4 & 10.5 & 18.5 & 22.5 \\
\method & \textbf{31.1} & \textbf{31.2} & \textbf{31.4} & \textbf{31.4} \\
\bottomrule
\end{tabular}
\end{table}

\begin{table}
\caption{Accuracy on LongBench v2 across token budgets (128--4096) at cluster size 16 for Qwen3 models (4B, 8B, 14B). CommunityKV accuracy improves with budget and reaches near dense parity at 4096 tokens across all model scales.}
\label{tab:main_eval}
\centering
\setlength{\tabcolsep}{3pt}
\begin{tabular}{@{}l@{\hspace{10pt}}lrrrrrr@{}}
\toprule
\textbf{Model} & \textbf{Method} & 128 & 256 & 512 & 1024 & 2048 & 4096 \\
\midrule
\multirow{3}{*}{\shortstack[l]{\textit{Qwen3-4B}\\\textit{(Dense: 27.4)}}}
& Quest      &  0.2 &  0.8 &  3.0 &  6.6 & 12.5 & 14.1 \\
& Multipole  & \textbf{25.8} & \textbf{26.2} & 25.6 & 25.8 & 25.6 & 25.8 \\
& \method   & 25.4 & 26.1 & \textbf{25.8} & \textbf{26.6} & \textbf{26.2} & \textbf{27.4} \\
\midrule
\multirow{3}{*}{\shortstack[l]{\textit{Qwen3-8B}\\\textit{(Dense: 31.4)}}}
& Quest      &  5.6 &  9.1 & 15.1 & 17.5 & 22.5 & 24.7 \\
& Multipole & \textbf{29.4} & \textbf{30.0} & 28.4 & 29.8 & 28.2 & 28.8 \\
& \method   & 28.5 & 28.8 & \textbf{29.5} & \textbf{30.1} & \textbf{29.9} & \textbf{31.1} \\
\midrule
\multirow{3}{*}{\shortstack[l]{\textit{Qwen3-14B}\\\textit{(Dense: 32.0)}}}
& Quest      &  9.3 & 18.1 & 21.5 & 24.7 & 29.0 & 30.6 \\
& Multipole  & \textbf{30.4} & 29.2 & 29.2 & 28.0 & 28.8 & 28.8 \\
& \method   & 30.2 & \textbf{30.0} & \textbf{30.5} & \textbf{30.7} & \textbf{29.7} & \textbf{32.0} \\
\bottomrule
\end{tabular}
\end{table}

\paragraph{Comparison between training-free retrieval methods}

We compare \method~against Quest and Multipole on LongBench v2 across the Qwen3 family (4B, 8B, 14B), sweeping token budgets from 128 to 4096 at cluster size 16 (Table~\ref{tab:main_eval}). \method~nearly achieves parity with exact FlashAttention-2 across all model scales at a 4096-token budget; for example, 32.0\% on Qwen3-14B, matching the dense upper bound exactly. Multipole performs strongly at restricted budgets (128--256 tokens) but plateaus as the budget increases, suggesting that $k$-means fails to capture the semantic structure required for complex reasoning. Quest scales with budget but trails dense by 1.4--13.3\% even at 4k tokens. \method~bridges this gap: its accuracy tends to rise with budget, translating additional retrieval capacity directly into better accuracy.

\paragraph{Multi-hop retrieval}

At a matched 4,096-token budget, CommunityKV achieves the strongest average among direct retrieval methods for both model families and the best or second-best average among all sparse baselines (Table~\ref{tab:babilong_comparison}). It slightly exceeds SnapKV on Llama-3.1-8B-Instruct (42.2\% versus 42.0\%) and trails it on Qwen3-8B (45.4\% versus 49.2\%).

\begin{table}
\caption{Accuracy on BABILong QA1--QA5 across model families and sparse-attention baselines. Results use 64k contexts and a matched 4,096-token budget. We compare CommunityKV with direct retrieval methods (TokenSelect, FreeKV, and SparQ) and eviction/compression methods (SnapKV, H2O, and StreamingLLM).}
\label{tab:babilong_comparison}
\centering
\setlength{\tabcolsep}{2pt}
\begin{tabular}{@{}lrrrrrr@{\hspace{10pt}}rrrrrr@{}}
\toprule
& \multicolumn{6}{c}{\textit{Qwen3-8B}}
& \multicolumn{6}{c}{\textit{Llama-3.1-8B-Instruct}} \\
\cmidrule(lr){2-7}\cmidrule(l){8-13}
\textbf{Method} & \textbf{QA1} & \textbf{QA2} & \textbf{QA3}
& \textbf{QA4} & \textbf{QA5} & \textbf{Avg.}
& \textbf{QA1} & \textbf{QA2} & \textbf{QA3}
& \textbf{QA4} & \textbf{QA5} & \textbf{Avg.} \\
\midrule
Dense        & 63 & 32 & 30 & 55 & 63 & 48.6 & 62 & 26 & 23 & 43 & 58 & 42.4 \\
CommunityKV  & 48 & 31 & 30 & 53 & 65 & 45.4 & 63 & 24 & 21 & 44 & 59 & \textbf{42.2} \\
TokenSelect  & 58 & 28 & 28 & 48 & 56 & 43.6 & 57 & 21 & 16 & 44 & 52 & 38.0 \\
FreeKV       & 52 & 28 & 23 & 48 & 58 & 41.8 & 58 & 23 & 22 & 47 & 57 & 41.4 \\
SparQ        & 49 & 32 & 17 & 51 & 57 & 41.2 & 59 & 15 & 15 & 44 & 54 & 37.4 \\
SnapKV       & 57 & 37 & 35 & 57 & 60 & 49.2 & 65 & 24 & 21 & 44 & 56 & 42.0 \\
H2O          & 20 & 18 & 22 & 45 & 68 & 34.6 & 15 & 10 & 19 & 50 & 55 & 29.8 \\
StreamingLLM & 7 & 12 & 35 & 47 & 72 & 34.6 & 10 & 14 & 28 & 50 & 64 & 33.2 \\
\bottomrule
\end{tabular}
\end{table}

\subsection{Computational efficiency}

We measure end-to-end throughput and peak GPU memory for 32,768-token generations at 64k and 128k input contexts across Qwen3-4B, 8B, and 14B on an NVIDIA H200 (Table~\ref{tab:full_request_efficiency}). Under the per-query-head configuration used for the primary accuracy evaluation, CommunityKV achieves $1.08\text{--}1.25\times$ full-request speedup. Graph and centroid storage is 52--82\% of KV-cache size, and total peak memory is 17--29\% above dense. Query-group aggregation reduces auxiliary storage to approximately 12--15\% of KV-cache size and peak-memory overhead to 0.5--8.5\%, while increasing full-request speedup to $1.39\text{--}1.71\times$. Throughput gains increase with context length for every model size. Query-group aggregation nearly matches per-query-head accuracy in the aggregation ablation (31.8\% versus 32.0\%; Table~\ref{tab:ablations}), making it the practical accuracy--systems tradeoff.

\begin{table}[t]
\caption{Generation of 32,768 tokens on an H200 with batch size 1. Timing spans prefill through final decode and includes all model and CommunityKV operations. Auxiliary storage and peak memory are reported in GiB. Auxiliary storage comprises graphs and centroids, with percentages relative to KV-cache size; peak-memory percentages are increases over dense attention.}
\label{tab:full_request_efficiency}
\centering
\setlength{\tabcolsep}{2pt}
\begin{tabular}{@{}l@{\hspace{6pt}}llrrrrrr@{}}
\toprule
\textbf{Model} & \textbf{Context} & \textbf{Graph Agg} & \textbf{Tok/s} & \textbf{Speedup}
& \textbf{Aux} & \textbf{Aux/KV}
& \textbf{Peak} & \textbf{$\Delta$ Peak} \\
\midrule
\multirow{6}{*}{\textit{Qwen3-4B}}
& 64k & Dense & 88.95 & 1.00$\times$ & & & 32.80 & \\
&      & Per-query head & 99.73 & 1.12$\times$ & 9.07 & 67.2\% & 39.69 & 21.0\% \\
&      & Query group & 133.86 & 1.50$\times$ & 2.03 & 15.0\% & 33.83 & 3.1\% \\
\cmidrule(l){2-9}
& 128k & Dense & 73.13 & 1.00$\times$ & & & 47.19 & \\
&      & Per-query head & 90.77 & 1.24$\times$ & 12.36 & 54.9\% & 60.65 & 28.5\% \\
&      & Query group & 125.37 & 1.71$\times$ & 2.69 & 12.0\% & 51.22 & 8.5\% \\
\midrule
\multirow{6}{*}{\textit{Qwen3-8B}}
& 64k & Dense & 75.86 & 1.00$\times$ & & & 40.81 & \\
&      & Per-query head & 83.97 & 1.10$\times$ & 8.57 & 63.5\% & 48.14 & 18.0\% \\
&      & Query group & 107.33 & 1.41$\times$ & 1.98 & 14.7\% & 42.72 & 4.7\% \\
\cmidrule(l){2-9}
& 128k & Dense & 63.49 & 1.00$\times$ & & & 61.95 & \\
&      & Per-query head & 79.08 & 1.25$\times$ & 11.63 & 51.7\% & 72.74 & 17.4\% \\
&      & Query group & 103.02 & 1.62$\times$ & 2.63 & 11.7\% & 63.86 & 3.1\% \\
\midrule
\multirow{6}{*}{\textit{Qwen3-14B}}
& 64k & Dense & 56.54 & 1.00$\times$ & & & 58.69 & \\
&      & Per-query head & 60.94 & 1.08$\times$ & 12.32 & 82.1\% & 69.52 & 18.5\% \\
&      & Query group & 78.95 & 1.39$\times$ & 2.30 & 15.3\% & 58.97 & 0.5\% \\
\cmidrule(l){2-9}
& 128k & Dense & 48.60 & 1.00$\times$ & & & 79.95 & \\
&      & Per-query head & 56.90 & 1.17$\times$ & 16.95 & 67.8\% & 97.16 & 21.5\% \\
&      & Query group & 75.50 & 1.55$\times$ & 3.15 & 12.6\% & 84.49 & 5.7\% \\
\bottomrule
\end{tabular}
\end{table}

\subsection{Ablations}
\label{sec:ablations}

We ablate the key design parameters of \method~on LongBench v2 with Qwen3-14B at a 4096-token budget, starting from the default configuration ($\kappa=8$, $\lambda=0.5$, graph per query head, 10 sink tokens excluded) (Table~\ref{tab:ablations}).
\paragraph{Graph sparsity ($\kappa$)} Accuracy peaks at $\kappa=8$ (32.0\%) and drops sharply at higher values (27.8\% at $\kappa=32$), suggesting that excessive connectivity obscures community boundaries. The lower accuracy with too few edges ($\kappa=2$, 30.3\%) similarly suggests graph fragmentation.

\paragraph{Direct vs.\ co-attention ($\lambda$)} Accuracy peaks at $\lambda=0.5$ (32.0\%) and degrades toward pure direct attention ($\lambda=1$, 28.5\%) or pure co-attention ($\lambda=0$, 28.8\%), confirming that neither signal is sufficient alone.

\paragraph{Centroid budget} By default, retrieval greedily fills the token budget with tokens from top-ranked communities. We test reserving part of the budget for additional community centroids as compressed summaries ($b_c \in \{0, 4, 16, 64, 256\}$). The greedy policy ($b_c=0$, 32.0\%) consistently outperforms centroid summaries (e.g., 28.6\% at $b_c=256$): full-token depth is more valuable than compressed breadth.

\paragraph{Graph aggregation} In grouped-query attention (GQA) architectures, query heads share KV heads. We test three granularities for graph construction: per-query-head (default), query-group (one graph per KV group), and layer-wise (one graph per layer). Per-head (32.0\%) and query-group (31.8\%) perform comparably, while layer-wise collapses to 21.5\%. Query-group aggregation is a viable memory-saving alternative.

\paragraph{Sink tokens} Excluding sink tokens from the graph and always retrieving them (32.0\%) outperforms including them in the partition (31.3\%). Sinks attract attention from nearly all positions, creating high-degree hubs that bridge unrelated communities.

\begin{table}
\caption{Ablation studies on LongBench v2 (Qwen3-14B, token budget=4096). The default configuration ($\kappa{=}8$, $\lambda{=}0.5$, per-head graphs, sinks excluded) is marked with \colorbox{gray!20}{gray shading} and achieves the best accuracy in every category.}
\label{tab:ablations}
\centering
\begin{tabular}{lccccc}
\toprule
\textbf{Graph Sparsity ($\kappa$)} & $\kappa=2$ & $\kappa=4$ & \colorbox{gray!20}{$\kappa=8$} & $\kappa=16$ & $\kappa=32$ \\
& 30.3 & 31.1 & \textbf{32.0} & 28.7 & 27.8 \\
\midrule
\textbf{Graph Weight ($\lambda$)} & $\lambda=0.0$ & $\lambda=0.25$ & \colorbox{gray!20}{$\lambda=0.50$} & $\lambda=0.75$ & $\lambda=1.0$ \\
 & 28.8 & 29.0 & \textbf{32.0} & 29.0 & 28.5 \\
\midrule
\textbf{Centroid Budget} & \colorbox{gray!20}{0} & 4 & 16 & 64 & 256 \\
 & \textbf{32.0} & 27.5 & 29.7 & 28.4 & 28.6 \\
\midrule
\textbf{Graph Aggregation} & \colorbox{gray!20}{No Agg.} & Query Group & Layer-Wise & -- & -- \\
 & \textbf{32.0} & 31.8 & 21.5 & -- & -- \\
\midrule
\textbf{Sink Tokens} & \colorbox{gray!20}{Excluded} & In-Graph & -- & -- & -- \\
& \textbf{32.0} & 31.3 & -- & -- & -- \\
\bottomrule
\end{tabular}
\end{table}

\paragraph{Isolating the contribution of Leiden}
We compare Leiden with two controls on the identical weighted graph: the connected-components control removes edges below a calibrated threshold and uses the surviving components as retrieval groups, while the top-$k$-neighborhood control seeds retrieval at the highest-scoring tokens and expands each seed by one graph hop to fill the budget. Both controls are calibrated to match CommunityKV's mean group size of 16. CommunityKV outperforms both at every budget and model scale (Table~\ref{tab:partitioning_controls}), isolating the benefit of Leiden from graph construction.

\begin{table}[t]
\caption{CommunityKV outperforms connected-components and top-$k$-neighborhood controls at every budget and model scale, highlighting the importance of Leiden partitioning.}
\label{tab:partitioning_controls}
\centering
\setlength{\tabcolsep}{3pt}
\begin{tabular}{@{}l@{\hspace{10pt}}lrrrrrr@{}}
\toprule
\textbf{Model} & \textbf{Method} & 128 & 256 & 512 & 1024 & 2048 & 4096 \\
\midrule
\multirow{3}{*}{\textit{Qwen3-4B}}
& Connected components & 1.1 & 2.7 & 3.5 & 7.3 & 12.1 & 20.3 \\
& Top-$k$ neighborhoods & 23.5 & 23.3 & 23.3 & 21.9 & 19.0 & 18.2 \\
& \method               & \textbf{25.4} & \textbf{26.1} & \textbf{25.8} & \textbf{26.6} & \textbf{26.2} & \textbf{27.4} \\
\midrule
\multirow{3}{*}{\textit{Qwen3-8B}}
& Connected components & 3.2 & 5.5 & 6.7 & 11.6 & 15.8 & 21.8 \\
& Top-$k$ neighborhoods & 27.6 & 28.4 & 27.6 & 26.4 & 28.8 & 26.2 \\
& \method               & \textbf{28.5} & \textbf{28.8} & \textbf{29.5} & \textbf{30.1} & \textbf{29.9} & \textbf{31.1} \\
\midrule
\multirow{3}{*}{\textit{Qwen3-14B}}
& Connected components & 0.3 & 1.8 & 3.5 & 9.7 & 19.2 & 27.0 \\
& Top-$k$ neighborhoods & 28.0 & 28.2 & 29.4 & 28.6 & 28.6 & 29.6 \\
& \method               & \textbf{30.2} & \textbf{30.0} & \textbf{30.5} & \textbf{30.7} & \textbf{29.7} & \textbf{32.0} \\
\bottomrule
\end{tabular}
\end{table}

\section{Conclusion}

We presented CommunityKV, a training-free framework that formulates sparse attention as dynamic community detection on a token graph induced from $QK^\top$ scores. Partitioning this graph with the Leiden algorithm during prefill yields communities that serve as the retrieval unit during decoding, with sub-linear $\mathcal{O}(\sqrt{n}\,d)$ cost per step. A constant-time local update rule assigns generated tokens to communities without global re-partitioning, enabling streaming decoding at fixed update cost per step. End-to-end throughput reaches $1.25\times$ per query head and $1.71\times$ with query-group aggregation, which nearly matches per-head accuracy while reducing auxiliary storage from 52--82\% to 12--15\% of KV-cache size.

\paragraph{Limitations and future work}

Our greedy community assignment freezes prior assignments, which may cause partition drift over very long generation horizons; periodically re-partitioning every $\sqrt{n}$ steps mitigates this at amortized $\mathcal{O}(\sqrt{n} \log n)$ cost per step, but adds latency in practice. The edge approximation during decoding is restricted to the retrieved set, so true top-$\kappa$ keys outside retrieved communities are missed, and retrieval recall affects the quality of incremental updates. Unstructured graphs and isolated-span tasks may also limit whole-community retrieval. These limitations suggest several natural extensions of the framework. First, incremental Leiden refinement~\cite{lin2026efficient}, performing local moves in the neighborhood of newly inserted tokens rather than single-node greedy assignment, could better maintain partition quality during decoding over longer horizons. Second, the hierarchical structure that Leiden naturally produces via its aggregation phase could enable $\mathcal{O}(d\log n)$ retrieval at longer contexts where even $\mathcal{O}(\sqrt{n})$ communities become expensive to score. Finally, sharing community structure across adjacent layers, whose attention patterns are often correlated, could reduce both graph memory and pipelining depth.

\newpage

\begin{ack}
\textbf{Funding.} Funding in direct support of this work was provided by Amazon.
\textbf{Competing interests.} The authors' Amazon affiliations are disclosed above; the authors declare no additional competing interests.
\end{ack}

\bibliographystyle{plainnat}
\bibliography{references}

\begin{thebibliography}{51}
\providecommand{\natexlab}[1]{#1}
\providecommand{\url}[1]{\texttt{#1}}
\expandafter\ifx\csname urlstyle\endcsname\relax
  \providecommand{\doi}[1]{doi: #1}\else
  \providecommand{\doi}{doi: \begingroup \urlstyle{rm}\Url}\fi

\bibitem[Bai et~al.(2025)Bai, Tu, Zhang, Peng, Wang, Lv, Cao, Xu, Hou, Dong,
  et~al.]{bai2025longbench}
Yushi Bai, Shangqing Tu, Jiajie Zhang, Hao Peng, Xiaozhi Wang, Xin Lv, Shulin
  Cao, Jiazheng Xu, Lei Hou, Yuxiao Dong, et~al.
\newblock Longbench v2: Towards deeper understanding and reasoning on realistic
  long-context multitasks.
\newblock In \emph{Proceedings of the 63rd Annual Meeting of the Association
  for Computational Linguistics (Volume 1: Long Papers)}, pages 3639--3664,
  2025.

\bibitem[Bairi et~al.(2023)Bairi, Sonwane, Kanade, et~al.]{bairi2023codeplan}
Ramakrishna Bairi, Atharv Sonwane, Aditya Kanade, et~al.
\newblock Codeplan: Repository-level coding using llms and planning.
\newblock \emph{arXiv preprint arXiv:2309.12499}, 2023.

\bibitem[Beltagy et~al.(2020)Beltagy, Peters, and Cohan]{beltagy2020longformer}
Iz~Beltagy, Matthew~E Peters, and Arman Cohan.
\newblock Longformer: The long-document transformer.
\newblock \emph{arXiv preprint arXiv:2004.05150}, 2020.

\bibitem[Blondel et~al.(2008)Blondel, Guillaume, Lambiotte, and
  Lefebvre]{blondel2008fast}
Vincent~D Blondel, Jean-Loup Guillaume, Renaud Lambiotte, and Etienne Lefebvre.
\newblock Fast unfolding of communities in large networks.
\newblock \emph{Journal of statistical mechanics: theory and experiment},
  2008\penalty0 (10):\penalty0 P10008, 2008.

\bibitem[Chen et~al.(2024)Chen, Sadhukhan, Ye, Zhou, Zhang, Nolte, Tian, Douze,
  Bottou, Jia, et~al.]{chen2024magicpig}
Zhuoming Chen, Ranajoy Sadhukhan, Zihao Ye, Yang Zhou, Jianyu Zhang, Niklas
  Nolte, Yuandong Tian, Matthijs Douze, Leon Bottou, Zhihao Jia, et~al.
\newblock Magicpig: Lsh sampling for efficient llm generation.
\newblock \emph{arXiv preprint arXiv:2410.16179}, 2024.

\bibitem[Child(2019)]{child2019generating}
Rewon Child.
\newblock Generating long sequences with sparse transformers.
\newblock \emph{arXiv preprint arXiv:1904.10509}, 2019.

\bibitem[Dao(2023)]{dao2023flashattention}
Tri Dao.
\newblock Flashattention-2: Faster attention with better parallelism and work
  partitioning.
\newblock \emph{arXiv preprint arXiv:2307.08691}, 2023.

\bibitem[{DeepSeek-AI}(2025)]{deepseekv32}
{DeepSeek-AI}.
\newblock Deepseek-v3.2: Pushing the frontier of open large language models.
\newblock \emph{arXiv preprint arXiv:2512.02556}, 2025.

\bibitem[Ding et~al.(2023)Ding, Ma, Dong, Zhang, Huang, Wang, Zheng, and
  Wei]{ding2023longnet}
Jiayu Ding, Shuming Ma, Li~Dong, Xingxing Zhang, Shaohan Huang, Wenhui Wang,
  Nanning Zheng, and Furu Wei.
\newblock Longnet: Scaling transformers to 1,000,000,000 tokens.
\newblock \emph{arXiv preprint arXiv:2307.02486}, 2023.

\bibitem[Erdogan et~al.(2025)]{erdogan2025plan}
Lutfi~Eren Erdogan et~al.
\newblock Plan-and-act: Improving planning of agents for long-horizon tasks.
\newblock \emph{arXiv preprint arXiv:2503.09572}, 2025.

\bibitem[Fender et~al.(2022)Fender, Rees, and Eaton]{fender2022rapids}
Alex Fender, Brad Rees, and Joe Eaton.
\newblock Rapids cugraph.
\newblock In \emph{Massive Graph Analytics}, pages 483--493. Chapman and
  Hall/CRC, 2022.

\bibitem[Fortunato and Barthelemy(2007)]{fortunato2007resolution}
Santo Fortunato and Marc Barthelemy.
\newblock Resolution limit in community detection.
\newblock \emph{Proceedings of the national academy of sciences}, 104\penalty0
  (1):\penalty0 36--41, 2007.

\bibitem[Ge et~al.(2023)Ge, Zhang, Liu, Zhang, Han, and Gao]{ge2023model}
Suyu Ge, Yunan Zhang, Liyuan Liu, Minjia Zhang, Jiawei Han, and Jianfeng Gao.
\newblock Model tells you what to discard: Adaptive kv cache compression for
  llms.
\newblock \emph{arXiv preprint arXiv:2310.01801}, 2023.

\bibitem[Grattafiori et~al.(2024)]{grattafiori2024llama}
Aaron Grattafiori et~al.
\newblock The llama 3 herd of models.
\newblock \emph{arXiv preprint arXiv:2407.21783}, 2024.

\bibitem[Han et~al.(2023)Han, Jayaram, Karbasi, Mirrokni, Woodruff, and
  Zandieh]{han2023hyperattention}
Insu Han, Rajesh Jayaram, Amin Karbasi, Vahab Mirrokni, David~P Woodruff, and
  Amir Zandieh.
\newblock Hyperattention: Long-context attention in near-linear time.
\newblock \emph{arXiv preprint arXiv:2310.05869}, 2023.

\bibitem[Hooper et~al.(2025)Hooper, Zhao, Manolache, Kim, Mahoney, Shao,
  Keutzer, and Gholami]{hooper2025multipole}
Coleman Hooper, Sebastian Zhao, Luca Manolache, Sehoon Kim, Michael~W Mahoney,
  Yakun~Sophia Shao, Kurt Keutzer, and Amir Gholami.
\newblock Multipole attention for efficient long context reasoning.
\newblock \emph{arXiv preprint arXiv:2506.13059}, 2025.

\bibitem[Hsieh et~al.(2024)Hsieh, Sun, Kriman, Acharya, Rekesh, Jia, Zhang, and
  Ginsburg]{hsieh2024ruler}
Cheng-Ping Hsieh, Simeng Sun, Samuel Kriman, Shantanu Acharya, Dima Rekesh, Fei
  Jia, Yang Zhang, and Boris Ginsburg.
\newblock Ruler: What's the real context size of your long-context language
  models?
\newblock \emph{arXiv preprint arXiv:2404.06654}, 2024.

\bibitem[Johnson et~al.(2021)Johnson, Douze, and J{\'e}gou]{johnson2019billion}
Jeff Johnson, Matthijs Douze, and Herv{\'e} J{\'e}gou.
\newblock Billion-scale similarity search with gpus.
\newblock \emph{IEEE Transactions on Big Data}, 7\penalty0 (3):\penalty0
  535--547, 2021.
\newblock \doi{10.1109/TBDATA.2019.2921572}.

\bibitem[Kitaev et~al.(2020)Kitaev, Kaiser, and Levskaya]{kitaev2020reformer}
Nikita Kitaev, {\L}ukasz Kaiser, and Anselm Levskaya.
\newblock Reformer: The efficient transformer.
\newblock \emph{arXiv preprint arXiv:2001.04451}, 2020.

\bibitem[Kuratov et~al.(2024)Kuratov, Bulatov, Anokhin, Rodkin, Sorokin,
  Sorokin, and Burtsev]{kuratov2024babilong}
Yuri Kuratov, Aydar Bulatov, Petr Anokhin, Ivan Rodkin, Dmitry Sorokin, Artyom
  Sorokin, and Mikhail Burtsev.
\newblock {BABILong}: Testing the limits of {LLMs} with long context
  reasoning-in-a-haystack.
\newblock In \emph{Advances in Neural Information Processing Systems
  (NeurIPS)}, 2024.

\bibitem[Lambiotte et~al.(2014)Lambiotte, Delvenne, and
  Barahona]{lambiotte2014random}
Renaud Lambiotte, Jean-Charles Delvenne, and Mauricio Barahona.
\newblock Random walks, {M}arkov processes and the multiscale modular
  organization of complex networks.
\newblock \emph{IEEE Transactions on Network Science and Engineering},
  1\penalty0 (2):\penalty0 76--90, 2014.

\bibitem[Li et~al.(2025)Li, Jin, and Zhang]{li2025graphkv}
Xuelin Li, Xiangqi Jin, and Linfeng Zhang.
\newblock Graphkv: Breaking the static selection paradigm with graph-based kv
  cache eviction.
\newblock In \emph{Proceedings of the 2025 Conference on Empirical Methods in
  Natural Language Processing}, pages 21910--21920, 2025.

\bibitem[Li et~al.(2024)Li, Huang, Yang, Venkitesh, Locatelli, Ye, Cai, Lewis,
  and Chen]{li2024snapkv}
Yuhong Li, Yingbing Huang, Bowen Yang, Bharat Venkitesh, Acyr Locatelli,
  Hanchen Ye, Tianle Cai, Patrick Lewis, and Deming Chen.
\newblock Snapkv: Llm knows what you are looking for before generation.
\newblock \emph{Advances in Neural Information Processing Systems},
  37:\penalty0 22947--22970, 2024.

\bibitem[Lin et~al.(2026)Lin, Xie, Fang, Hu, Hu, and Cheng]{lin2026efficient}
Chunxu Lin, Yumao Xie, Yixiang Fang, Yongmin Hu, Yingqian Hu, and Chen Cheng.
\newblock Efficient maintenance of leiden communities in large dynamic graphs.
\newblock \emph{arXiv preprint arXiv:2601.08554}, 2026.

\bibitem[Liu et~al.(2025)Liu, Li, Ning, Lin, Yao, Ke, Guo, and
  Zhao]{liu2025freekv}
Guangda Liu, Chengwei Li, Zhenyu Ning, Jing Lin, Yiwu Yao, Danning Ke, Minyi
  Guo, and Jieru Zhao.
\newblock Freekv: Boosting kv cache retrieval for efficient llm inference.
\newblock \emph{arXiv preprint arXiv:2505.13109}, 2025.

\bibitem[Liu et~al.(2023)Liu, Desai, Liao, Wang, Xie, Xu, Kyrillidis, and
  Shrivastava]{liu2023scissorhands}
Zichang Liu, Aditya Desai, Fangshuo Liao, Weitao Wang, Victor Xie, Zhaozhuo Xu,
  Anastasios Kyrillidis, and Anshumali Shrivastava.
\newblock Scissorhands: Exploiting the persistence of importance hypothesis for
  llm kv cache compression at test time.
\newblock \emph{Advances in Neural Information Processing Systems},
  36:\penalty0 52342--52364, 2023.

\bibitem[Lu et~al.(2025)Lu, Jiang, Liu, Du, Jiang, Hong, Liu, He, Yuan, Wang,
  et~al.]{lu2025moba}
Enzhe Lu, Zhejun Jiang, Jingyuan Liu, Yulun Du, Tao Jiang, Chao Hong, Shaowei
  Liu, Weiran He, Enming Yuan, Yuzhi Wang, et~al.
\newblock Moba: Mixture of block attention for long-context llms.
\newblock \emph{arXiv preprint arXiv:2502.13189}, 2025.

\bibitem[Nawrot et~al.(2025)Nawrot, Li, Huang, Ruder, Marchisio, and
  Ponti]{nawrot2025sparse}
Piotr Nawrot, Robert Li, Renjie Huang, Sebastian Ruder, Kelly Marchisio, and
  Edoardo~M Ponti.
\newblock The sparse frontier: Sparse attention trade-offs in transformer llms.
\newblock \emph{arXiv preprint arXiv:2504.17768}, 2025.

\bibitem[Newman(2006)]{newman2006modularity}
Mark~EJ Newman.
\newblock Modularity and community structure in networks.
\newblock \emph{Proceedings of the National Academy of Sciences}, 103\penalty0
  (23):\penalty0 8577--8582, 2006.

\bibitem[Peng et~al.(2023)Peng, Quesnelle, Fan, and Shippole]{peng2023yarn}
Bowen Peng, Jeffrey Quesnelle, Honglu Fan, and Enrico Shippole.
\newblock Yarn: Efficient context window extension of large language models.
\newblock \emph{arXiv preprint arXiv:2309.00071}, 2023.

\bibitem[Rae et~al.(2020)Rae, Potapenko, Jayakumar, Hillier, and
  Lillicrap]{rae2020compressive}
Jack~W. Rae, Anna Potapenko, Siddhant~M. Jayakumar, Chloe Hillier, and
  Timothy~P. Lillicrap.
\newblock Compressive transformers for long-range sequence modelling.
\newblock In \emph{International Conference on Learning Representations}, 2020.

\bibitem[Rando et~al.(2025)Rando, Romani, Sampieri, Franco, Yang, Kyuragi,
  Galasso, and Hashimoto]{rando2025longcodebench}
Stefano Rando, Luca Romani, Alessio Sampieri, Luca Franco, John Yang, Yuta
  Kyuragi, Fabio Galasso, and Tatsunori Hashimoto.
\newblock Longcodebench: Evaluating coding llms at 1m context windows.
\newblock \emph{arXiv preprint arXiv:2505.07897}, 2025.

\bibitem[Ribar et~al.(2024)Ribar, Chelombiev, Hudlass-Galley, Blake, Luschi,
  and Orr]{ribar2024sparq}
Luka Ribar, Ivan Chelombiev, Luke Hudlass-Galley, Charlie Blake, Carlo Luschi,
  and Douglas Orr.
\newblock Sparq attention: Bandwidth-efficient llm inference.
\newblock \emph{Proceedings of the 41st International Conference on Machine
  Learning}, 2024.

\bibitem[Roy et~al.(2021)Roy, Saffar, Vaswani, and Grangier]{roy2021efficient}
Aurko Roy, Mohammad Saffar, Ashish Vaswani, and David Grangier.
\newblock Efficient content-based sparse attention with routing transformers.
\newblock \emph{Transactions of the Association for Computational Linguistics},
  9:\penalty0 53--68, 2021.

\bibitem[Sahu(2024)]{sahu2024heuristic}
Subhajit Sahu.
\newblock Heuristic-based dynamic leiden algorithm for efficient tracking of
  communities on evolving graphs.
\newblock \emph{arXiv preprint arXiv:2410.15451}, 2024.

\bibitem[Sahu et~al.(2024)Sahu, Kothapalli, and Banerjee]{sahu2024fast}
Subhajit Sahu, Kishore Kothapalli, and Dip~Sankar Banerjee.
\newblock Fast leiden algorithm for community detection in shared memory
  setting.
\newblock In \emph{Proceedings of the 53rd International Conference on Parallel
  Processing}, pages 11--20, 2024.

\bibitem[Tang et~al.(2024)Tang, Zhao, Zhu, Xiao, Kasikci, and
  Han]{tang2024quest}
Jiaming Tang, Yilong Zhao, Kan Zhu, Guangxuan Xiao, Baris Kasikci, and Song
  Han.
\newblock Quest: Query-aware sparsity for efficient long-context llm inference.
\newblock \emph{arXiv preprint arXiv:2406.10774}, 2024.

\bibitem[Tay et~al.(2022)Tay, Dehghani, Bahri, and Metzler]{tay2022efficient}
Yi~Tay, Mostafa Dehghani, Dara Bahri, and Donald Metzler.
\newblock Efficient transformers: A survey.
\newblock \emph{ACM Computing Surveys (CSUR)}, 55\penalty0 (6):\penalty0 1--28,
  2022.
\newblock \doi{10.1145/3530811}.

\bibitem[Traag et~al.(2019)Traag, Waltman, and Van~Eck]{traag2019louvain}
Vincent~A Traag, Ludo Waltman, and Nees~Jan Van~Eck.
\newblock From louvain to leiden: guaranteeing well-connected communities.
\newblock \emph{Scientific reports}, 9\penalty0 (1):\penalty0 1--12, 2019.

\bibitem[Vaswani et~al.(2017)Vaswani, Shazeer, Parmar, Uszkoreit, Jones, Gomez,
  Kaiser, and Polosukhin]{vaswani2017attention}
Ashish Vaswani, Noam Shazeer, Niki Parmar, Jakob Uszkoreit, Llion Jones,
  Aidan~N Gomez, {\L}ukasz Kaiser, and Illia Polosukhin.
\newblock Attention is all you need.
\newblock \emph{Advances in neural information processing systems}, 30, 2017.

\bibitem[Vyas et~al.(2020)Vyas, Katharopoulos, and Fleuret]{vyas2020fast}
Apoorv Vyas, Angelos Katharopoulos, and Fran{\c{c}}ois Fleuret.
\newblock Fast transformers with clustered attention.
\newblock \emph{Advances in Neural Information Processing Systems},
  33:\penalty0 21665--21674, 2020.

\bibitem[Wang et~al.(2025)Wang, Shi, Tan, Qin, Wang, Zhang, Nambi, Ganu, and
  Wang]{wang2025multimodal}
Hengyi Wang, Haizhou Shi, Shiwei Tan, Weiyi Qin, Wenyuan Wang, Tunyu Zhang,
  Akshay Nambi, Tanuja Ganu, and Hao Wang.
\newblock Multimodal needle in a haystack: Benchmarking long-context capability
  of multimodal large language models.
\newblock In \emph{Proceedings of the 2025 Conference of the Nations of the
  Americas Chapter of the Association for Computational Linguistics: Human
  Language Technologies (Volume 1: Long Papers)}, pages 3221--3241, 2025.

\bibitem[Wang et~al.(2021)Wang, Zhou, Gan, Chen, Fang, Sun, Cheng, and
  Liu]{wang2021cluster}
Shuohang Wang, Luowei Zhou, Zhe Gan, Yen-Chun Chen, Yuwei Fang, Siqi Sun,
  Yu~Cheng, and Jingjing Liu.
\newblock Cluster-former: Clustering-based sparse transformer for question
  answering.
\newblock In \emph{Findings of the association for computational linguistics:
  ACL-IJCNLP 2021}, pages 3958--3968, 2021.

\bibitem[Wu et~al.(2025)Wu, Pan, Wang, Chen, Bai, Wang, Fu, Wang, and
  Xiong]{wu2025tokenselect}
Wei Wu, Zhuoshi Pan, Chao Wang, Liyi Chen, Yunchu Bai, Tianfu Wang, Kun Fu,
  Zheng Wang, and Hui Xiong.
\newblock Tokenselect: Efficient long-context inference and length
  extrapolation for llms via dynamic token-level kv cache selection.
\newblock \emph{Proceedings of the 2025 Conference on Empirical Methods in
  Natural Language Processing}, 2025.

\bibitem[Xiao et~al.(2023)Xiao, Tian, Chen, Han, and Lewis]{xiao2023efficient}
Guangxuan Xiao, Yuandong Tian, Beidi Chen, Song Han, and Mike Lewis.
\newblock Efficient streaming language models with attention sinks.
\newblock \emph{arXiv preprint arXiv:2309.17453}, 2023.

\bibitem[Xiao et~al.(2024)Xiao, Tian, Chen, Han, and
  Lewis]{xiao2024efficientstreaminglanguagemodels}
Guangxuan Xiao, Yuandong Tian, Beidi Chen, Song Han, and Mike Lewis.
\newblock Efficient streaming language models with attention sinks, 2024.
\newblock URL \url{https://arxiv.org/abs/2309.17453}.

\bibitem[Yang et~al.(2025)Yang, Li, Yang, Zhang, Hui, Zheng, Yu, Gao, Huang,
  Lv, et~al.]{yang2025qwen3}
An~Yang, Anfeng Li, Baosong Yang, Beichen Zhang, Binyuan Hui, Bo~Zheng, Bowen
  Yu, Chang Gao, Chengen Huang, Chenxu Lv, et~al.
\newblock Qwen3 technical report.
\newblock \emph{arXiv preprint arXiv:2505.09388}, 2025.

\bibitem[Yuan et~al.(2025)Yuan, Gao, Dai, Luo, Zhao, Zhang, Xie, Wei, Wang,
  Xiao, et~al.]{yuan2025native}
Jingyang Yuan, Huazuo Gao, Damai Dai, Junyu Luo, Liang Zhao, Zhengyan Zhang,
  Zhenda Xie, Yuxing Wei, Lean Wang, Zhiping Xiao, et~al.
\newblock Native sparse attention: Hardware-aligned and natively trainable
  sparse attention.
\newblock In \emph{Proceedings of the 63rd Annual Meeting of the Association
  for Computational Linguistics (Volume 1: Long Papers)}, pages 23078--23097,
  2025.

\bibitem[Zaheer et~al.(2020)Zaheer, Guruganesh, Dubey, Ainslie, Alberti,
  Ontanon, Pham, Ravula, Wang, Yang, et~al.]{zaheer2020big}
Manzil Zaheer, Guru Guruganesh, Kumar~Avinava Dubey, Joshua Ainslie, Chris
  Alberti, Santiago Ontanon, Philip Pham, Anirudh Ravula, Qifan Wang, Li~Yang,
  et~al.
\newblock Big bird: Transformers for longer sequences.
\newblock \emph{Advances in neural information processing systems},
  33:\penalty0 17283--17297, 2020.

\bibitem[Zandieh et~al.(2023)Zandieh, Han, Daliri, and
  Karbasi]{zandieh2023kdeformer}
Amir Zandieh, Insu Han, Majid Daliri, and Amin Karbasi.
\newblock Kdeformer: Accelerating transformers via kernel density estimation.
\newblock In \emph{International Conference on Machine Learning}, pages
  40605--40623. PMLR, 2023.

\bibitem[Zhang et~al.(2023)Zhang, Sheng, Zhou, Chen, Zheng, Cai, Song, Tian,
  R{\'e}, Barrett, et~al.]{zhang2023h2o}
Zhenyu Zhang, Ying Sheng, Tianyi Zhou, Tianlong Chen, Lianmin Zheng, Ruisi Cai,
  Zhao Song, Yuandong Tian, Christopher R{\'e}, Clark Barrett, et~al.
\newblock H2o: Heavy-hitter oracle for efficient generative inference of large
  language models.
\newblock \emph{Advances in Neural Information Processing Systems},
  36:\penalty0 34661--34710, 2023.

\end{thebibliography}

\newpage
\appendix

\section{Community scaling and resolution parameter}
\label{app:community_scaling}

\begin{figure}[!h]
\centering
\includegraphics[width=.9\linewidth]{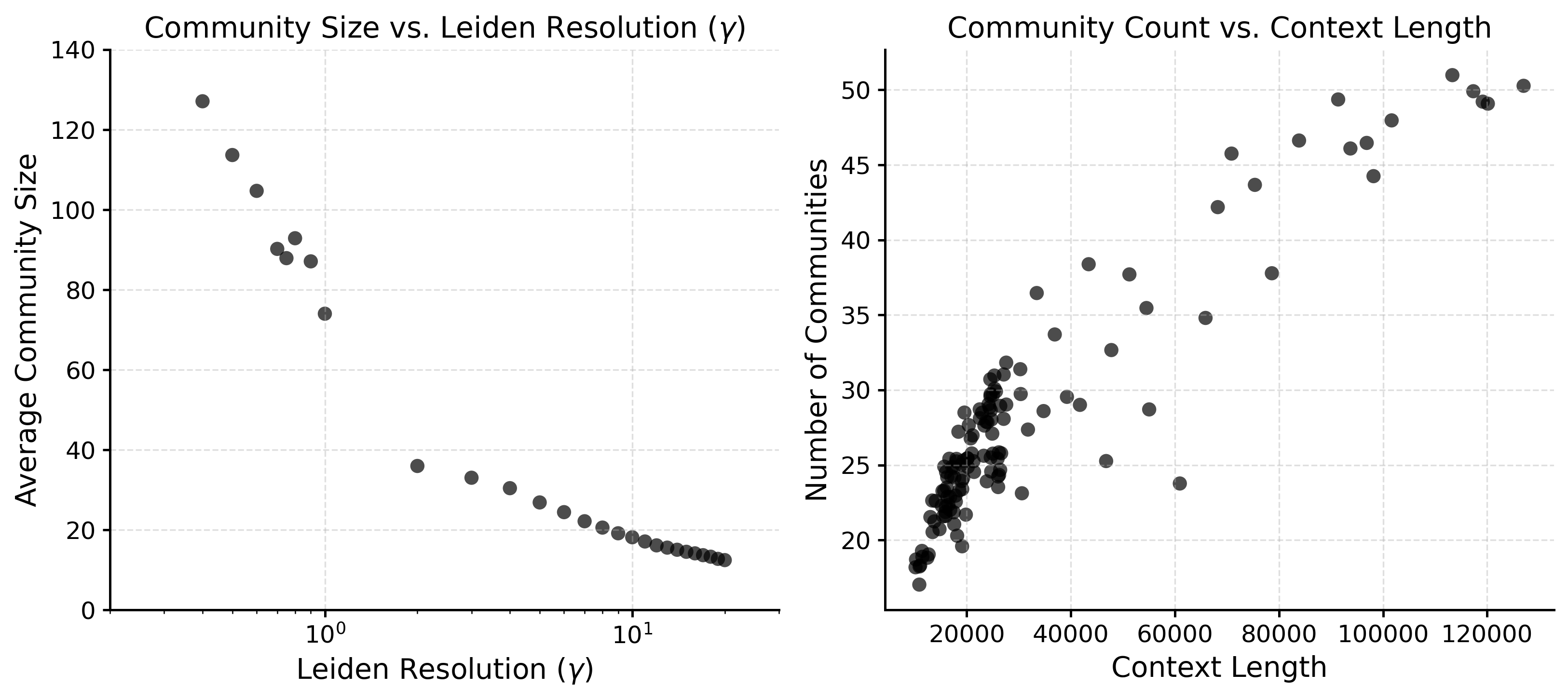}
\caption{Empirical analysis of graph partition properties on LongBench v2 (Qwen3-14B). \textbf{Left:} Average community size as a function of the resolution parameter $\gamma$. \textbf{Right:} Number of communities vs.\ context length, confirming $\mathcal{O}(\sqrt{n})$ sub-linear scaling.}
\label{fig:community_scaling}
\end{figure}

We empirically verify two properties of the graph partition on LongBench v2 using Qwen3-14B (Figure~\ref{fig:community_scaling}).

\subsection{Resolution parameter $\gamma$}

The left panel shows the relationship between the Leiden resolution parameter $\gamma$ and the average community size. Increasing $\gamma$ produces smaller, denser communities, providing a direct control over retrieval granularity. In our experiments, we tune $\gamma$ to match a cluster size of 16 for fair comparison with each baseline (Section~\ref{sec:results_long_context}). On Qwen3-8B at a 4,096-token budget, scaling the selected $\gamma$ over a $64\times$ range changes the predicted answer on at most 2 examples in a fixed set of 30, despite a $2.5\times$ change in mean community size.

\subsection{Sub-linear community scaling}

The right panel shows the number of communities as a function of sequence length. The empirical scaling closely follows $\mathcal{O}(\sqrt{n})$, consistent with the resolution limit of modularity optimization~\cite{fortunato2007resolution}. At 100k tokens, the graph partitions into approximately 50 communities. This sub-linear growth is what enables $\mathcal{O}(\sqrt{n}\,d)$ retrieval cost: scoring a query against $\mathcal{O}(\sqrt{n})$ centroids is substantially cheaper than the $\mathcal{O}(nd)$ cost of scoring against all tokens.

\section{Long-generation stability}
\label{app:long_generation_stability}

\subsection{Incremental partition drift}
\label{app:drift}

We stress-test the stability of our incremental update rule by tracking two metrics over a 4096-token generation horizon on LongBench v2 with Qwen3-14B, using the default configuration ($\kappa=8$, $\lambda=0.5$, 4096-token retrieval budget). No global re-partitioning is performed during generation.

We monitor (1) graph modularity $Q$ (Eq.~\ref{eq:modularity_score}), which measures how well the current partition reflects the evolving graph structure, and (2) attention mass recall $\tilde{\mu}_\kappa = \sum_{j \in \mathcal{R}_i} \alpha_{ij}$, the fraction of the full attention distribution captured by the retrieved set at each decoding step.

\begin{figure}
\centering
\includegraphics[width=.9\linewidth]{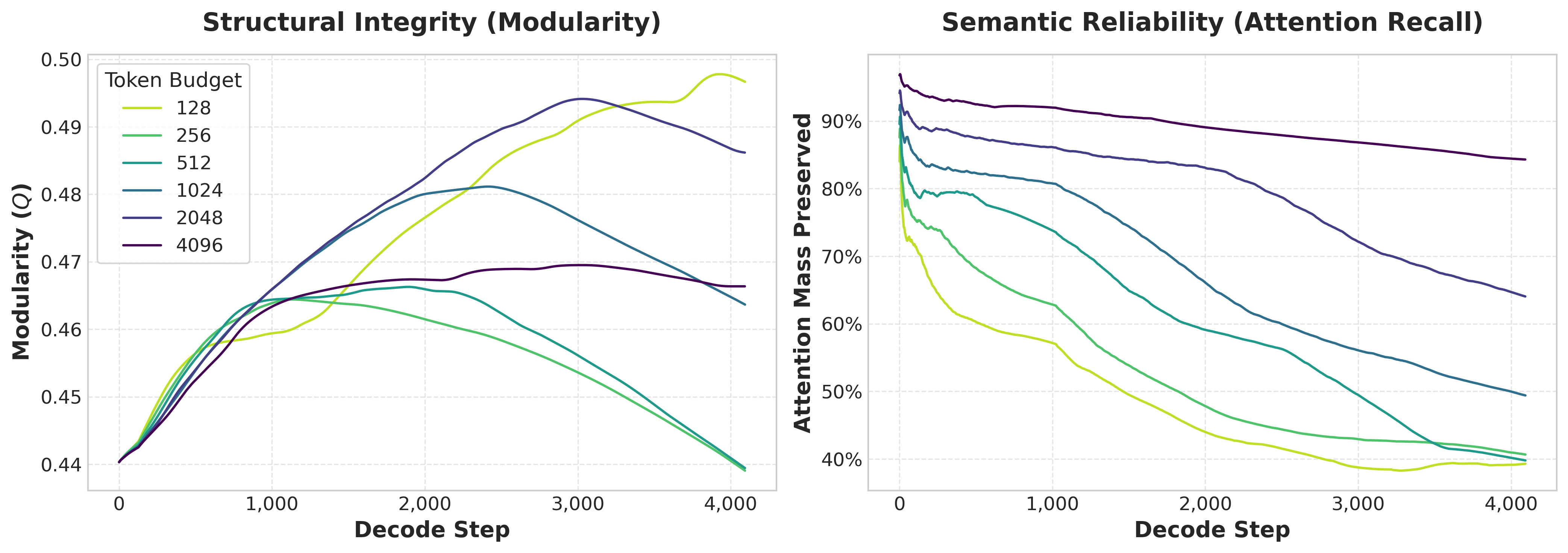}
\caption{Partition stability over a 4096-token generation horizon (Qwen3-14B, LongBench v2). \textbf{Left:} Graph modularity $Q$ remains stable and increases slightly as new tokens reinforce community structure. \textbf{Right:} Attention mass recall $\tilde{\mu}_\kappa$ at varying retrieval budgets. At the default 4096-token budget, recall stays above 85\% throughout generation.}
\label{fig:drift}
\end{figure}

Figure~\ref{fig:drift} shows both metrics as a function of generated tokens. Modularity remains in $[0.44, 0.50]$ throughout the 4096-token horizon, indicating that the greedy assignment rule continues to improve partition quality as new tokens reinforce existing community structure. Attention mass recall decays gracefully: at the 4096-token retrieval budget, recall remains above 85\% after 4096 generated tokens. Smaller budgets (128--512 tokens) exhibit faster decay, as expected, but do not collapse catastrophically.

We evaluate Qwen3-8B on 65,536-token prompts from PG-19 and Python using the default per-query-head configuration, a 4,096-token retrieval budget, $\kappa=8$, $\lambda=0.5$, and 10 sink tokens. We compare purely incremental graph updates with periodic global repartitioning every 512 generated tokens (Refresh-512).

\subsection{Fixed continuations}
\label{app:fixed_continuations}

We teacher-force the same 16,384-token held-out continuation through every method, isolating accumulated approximation error from differences in generated histories. Results average five prompts per domain and are measured over the terminal 512-token window at each horizon. We report negative log-likelihood (NLL), KL divergence from dense, and dense top-1 agreement.

\begin{table}[!ht]
\caption{Fixed-continuation stability over generation length. Lower NLL and KL are better; higher top-1 agreement is better.}
\label{tab:fixed_continuation_stability}
\centering
\setlength{\tabcolsep}{3pt}
\begin{tabular}{@{}llrrrrrrr@{}}
\toprule
& & \multicolumn{3}{c}{NLL} & \multicolumn{2}{c}{KL}
& \multicolumn{2}{c}{Top-1 agreement} \\
\cmidrule(lr){3-5}\cmidrule(lr){6-7}\cmidrule(l){8-9}
\textbf{Domain} & \textbf{Horizon}
& \textbf{Dense} & \textbf{Incr.} & \textbf{Refresh}
& \textbf{Incr.} & \textbf{Refresh}
& \textbf{Incr.} & \textbf{Refresh} \\
\midrule
\multirow{4}{*}{Python}
& 2k  & 0.7045 & 0.7534 & 0.7375 & 0.0600 & 0.0311 & 94.34\% & 95.63\% \\
& 4k  & 0.4952 & 0.5478 & 0.5288 & 0.0694 & 0.0369 & 94.53\% & 95.59\% \\
& 8k  & 0.7264 & 0.8470 & 0.7729 & 0.1372 & 0.0418 & 92.30\% & 95.59\% \\
& 16k & 0.7042 & 0.8297 & 0.7414 & 0.2510 & 0.0702 & 87.50\% & 94.30\% \\
\midrule
\multirow{4}{*}{PG-19}
& 2k  & 2.5129 & 2.6743 & 2.6158 & 0.1525 & 0.1144 & 82.81\% & 86.48\% \\
& 4k  & 2.3151 & 2.4675 & 2.3825 & 0.1991 & 0.0857 & 79.30\% & 86.80\% \\
& 8k  & 2.4834 & 2.7159 & 2.5853 & 0.2880 & 0.1109 & 75.23\% & 84.61\% \\
& 16k & 2.6481 & 3.0052 & 2.7873 & 0.4401 & 0.1453 & 68.05\% & 82.81\% \\
\bottomrule
\end{tabular}
\end{table}

\subsection{Sampled generation}

We generate 8,192 tokens at temperature 0.7 and top-$p=0.95$ using 10 prompts per domain and two sampling seeds. Because approximate methods produce different histories, we replay each generated history through dense attention and the generating method, then compare their distributions over the terminal 512-token window at 2k, 4k, and 8k tokens. NLL is reported only for the fixed-continuation experiment, where every method shares the same target tokens.

\begin{table}[!ht]
\caption{Sampled-generation stability over generation length. Values compare dense attention with each method on the same generated history.}
\label{tab:sampled_generation_stability}
\centering
\setlength{\tabcolsep}{5pt}
\begin{tabular}{@{}llrrrr@{}}
\toprule
& & \multicolumn{2}{c}{KL}
& \multicolumn{2}{c}{Top-1 agreement} \\
\cmidrule(lr){3-4}\cmidrule(l){5-6}
\textbf{Domain} & \textbf{Horizon}
& \textbf{Incr.} & \textbf{Refresh}
& \textbf{Incr.} & \textbf{Refresh} \\
\midrule
\multirow{3}{*}{PG-19}
& 2k & 0.0519 & 0.0521 & 93.96\% & 93.91\% \\
& 4k & 0.0083 & 0.0196 & 99.34\% & 97.87\% \\
& 8k & 0.0045 & 0.0097 & 99.75\% & 99.13\% \\
\midrule
\multirow{3}{*}{Python}
& 2k & 0.0290 & 0.0150 & 98.13\% & 99.01\% \\
& 4k & 0.0146 & 0.0105 & 99.21\% & 99.11\% \\
& 8k & 0.0163 & 0.0053 & 99.03\% & 99.61\% \\
\bottomrule
\end{tabular}
\end{table}

\section{Implementation and systems analysis}
\label{app:impl}

This appendix provides the full complexity derivation for fused top-$\kappa$ selection, the pipelining schedule that masks Leiden latency, and the memory accounting for per-head graphs.

\subsection{Fused top-$\kappa$ selection}
\label{app:fused_topk}

We construct the initial sparse graph by identifying the top-$\kappa$ keys for all $n$ queries simultaneously, fusing this selection directly into the FlashAttention kernel with a WarpSelect-style register-resident selector~\cite{johnson2019billion}. As the kernel computes scores for each $b_q \times b_k$ tile of queries and keys, each GPU thread maintains a sorted queue of $\kappa$ candidates using compare-and-insert operations. Each candidate contains a pre-softmax scaled QK logit and its key index. Block-level candidates are accumulated with warp shuffles to emit the row's top-$\kappa$ indices and corresponding weights, which are normalized using the row's final softmax statistics. The total complexity aggregates two costs:
\begin{enumerate}
\item \textbf{Block-level sorting.} Every score in the causal attention matrix is tested against a queue threshold, requiring $\mathcal{O}(n^2)$ work. In the worst case, each score triggers an insertion requiring at most $\kappa$ compare-and-insert operations, giving $\mathcal{O}(\kappa n^2)$ work.
\item \textbf{Global top-$\kappa$ reduction.} Producing $\kappa$ outputs requires a fixed-depth shuffle reduction for each selected candidate. Across all query rows, this costs $\mathcal{O}(\kappa n)$.
\end{enumerate}

Combining these terms, the total graph construction worst case complexity is:
\begin{equation}
\mathcal{O}\!\left(\kappa n^2+n\kappa\right)
=\mathcal{O}(\kappa n^2).
\end{equation}
Because $\kappa$ is fixed in our experiments, this scales as $\mathcal{O}(n^2)$, matching the underlying causal prefill computation and adding no asymptotic overhead.

This bound is pessimistic because an insertion occurs only when a score exceeds the queue's current $\kappa$-th largest value. Total CommunityKV prefill overheads are $7.8\%$ for Qwen3-4B, $7.7\%$ for Qwen3-8B, and $7.2\%$ for Qwen3-14B on 131k-token inputs (Table~\ref{tab:latency_breakdown}).

\subsection{Pipelined graph partitioning}
\label{app:pipelined_partitioning}

\begin{figure}[h]
\centering
\includegraphics[width=\linewidth]{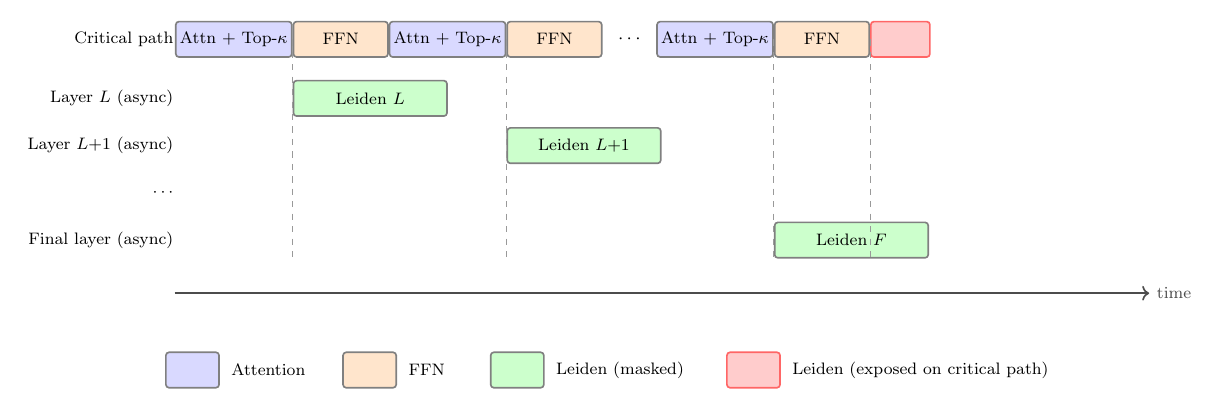}
\caption{Pipeline schedule for graph partitioning during prefill. Leiden partitioning for each layer is dispatched asynchronously after its attention scores are computed, overlapping with subsequent layer execution. The red segment indicates the residual Leiden latency that remains exposed on the critical path when the final layer's partitioning has not completed by the time it is needed.}
\label{fig:pipeline}
\end{figure}

Naively, Leiden partitioning would extend the prefill critical path by the sum of per-layer partition costs. We instead dispatch each layer's partitioning to asynchronous workers immediately after its attention scores are computed, allowing it to run in parallel with the FFN of the same layer and the attention of subsequent layers (Figure~\ref{fig:pipeline}). Because the FFN and following layers' attention together exceed the per-layer Leiden cost, all layers except the last have their partition cost fully masked.

The exposed latency is therefore the cost of partitioning a single layer's graphs, which we measure at 0.37--0.45\,s across the Qwen3 family on an NVIDIA H100 (Table~\ref{tab:latency_breakdown}). This corresponds to 2--3\% of total prefill wall time on a 130k-token input.

\subsection{Component-level latency breakdown}
\label{app:latency_breakdown}

\begin{table}[!ht]
\caption{Latency comparison between FlashAttention-2 (Dense) and CommunityKV using Hugging Face's \texttt{DynamicCache} on an NVIDIA H100 80GB at 131k context. Left: total prefill wall time with breakdown into fused top-$\kappa$ selection and exposed Leiden partitioning latency. Right: average per-step decoding wall time (128 generated tokens) with breakdown into token retrieval and sparse attention. Only attention-related kernels are reported; the remaining per-step time is MLP and layer normalization.}
\label{tab:latency_breakdown}
\centering
\setlength{\tabcolsep}{4pt}
\begin{tabular}{@{}l r r r @{\hskip 20pt} l r r r@{}}
\toprule
\multicolumn{4}{c}{\textbf{Prefill -- Total Wall Time (s)}} & \multicolumn{4}{c}{\textbf{Decoding -- Avg Per-Step Wall Time (ms)}} \\
\cmidrule(r){1-4} \cmidrule(l){5-8}
& \textbf{4B} & \textbf{8B} & \textbf{14B} & & \textbf{4B} & \textbf{8B} & \textbf{14B} \\
\midrule
Dense & 16.5 & 18.2 & 26.3 & Dense & 63 & 64 & 71 \\
\method & 17.8 & 19.6 & 28.2 & \method & 37 & 34 & 51 \\
\hspace{2mm} Top-$\kappa$ Sel. & 0.94 & 1.09 & 1.49 & \hspace{2mm} Retrieval & 5.3 & 4.7 & 7.2 \\
\hspace{2mm} Leiden & 0.37 & 0.45 & 0.42 & \hspace{2mm} Attention & 6.1 & 5.5 & 8.2 \\
\midrule
\textit{Overhead} & +7.8\% & +7.7\% & +7.2\% & \textbf{\textit{Speedup}} & \textbf{1.69$\times$} & \textbf{1.86$\times$} & \textbf{1.39$\times$} \\
\bottomrule
\end{tabular}
\end{table}

\subsection{Memory footprint}

CommunityKV maintains an independent attention graph per query head (or per KV group or per layer under aggregation). For modern multi-layer, multi-head models, the aggregate footprint of these graphs warrants explicit accounting.

\paragraph{Per-graph footprint} Each graph stores at most $|E| \le 2n\kappa(1+\kappa)$ edges (Section~\ref{sec:graph_construction}). Storing edge weights in bfloat16 with 32-bit indices yields a per-graph cost of $8 |E|$ bytes. At $n = 100$k tokens with $\kappa = 8$, we empirically observe $|E| \approx 1.15$M edges and a per-graph footprint of ${\sim}9$\,MB; the observed footprint is ${<}20$\,MB at $\kappa = 8$ across all sequence lengths in our evaluation.

\paragraph{Aggregate footprint} Without aggregation, a model with $L$ layers and $H$ query heads maintains $L \cdot H$ independent graphs. For Qwen3-14B ($L=40$, $H=40$, with 8 KV heads per layer), this is 1600 per-head graphs. Under query-group aggregation (one graph per KV group, ablated in Section~\ref{sec:ablations}), this reduces to $L \cdot H_{\text{KV}} = 320$ graphs. At 100k tokens with $\kappa=8$ in bfloat16, the aggregate footprint is ${<}15$\,GB without aggregation and ${<}3$\,GB with query-group aggregation, compared to ${\sim}16$\,GB for the dense KV cache at the same context length and the 28\,GB Qwen3-14B model weights in bfloat16. Graph memory under query-group aggregation is therefore a small fraction of total model state.

\paragraph{Centroid storage} Community centroids are stored in $d$-dimensional bfloat16. With $\mathcal{O}(\sqrt{n})$ communities per graph (Appendix~\ref{app:community_scaling}), centroid storage is $\mathcal{O}(\sqrt{n}\,d)$ per graph and is dominated by the edge storage above at all evaluated context lengths.

\paragraph{Comparison to dense KV cache} The dense KV cache scales as $2 L \cdot H_{\text{KV}} \cdot n \cdot d_{\text{head}}$ in bfloat16. For Qwen3-14B at 100k context, this is approximately 16\,GB, roughly an order of magnitude larger than the graph footprint under query-group aggregation. Because CommunityKV preserves the full KV cache (graphs serve as a routing index, not a replacement), graph memory is purely additive but small relative to the cache it indexes.

\section{Expanded baseline results}
\label{app:expanded_baselines}

\subsection{LongBench v2}

\begin{table}[!ht]
\caption{Accuracy on LongBench v2 across matched token budgets for direct retrieval, eviction/compression, and CommunityKV methods.}
\label{tab:longbench_all_baselines}
\centering
\setlength{\tabcolsep}{4pt}
\begin{tabular}{@{}lrrrrrr@{}}
\toprule
\textbf{Method} & \textbf{128} & \textbf{256} & \textbf{512}
& \textbf{1024} & \textbf{2048} & \textbf{4096} \\
\midrule
\multicolumn{7}{c}{\textit{Qwen3-4B} (Dense: 27.4)} \\
\midrule
Quest        & 0.2 & 0.8 & 3.0 & 6.6 & 12.5 & 14.1 \\
Multipole    & 25.8 & 26.2 & 25.6 & 25.8 & 25.6 & 25.8 \\
TokenSelect  & 26.1 & 26.1 & 25.9 & 25.7 & 25.9 & 25.3 \\
FreeKV       & 22.6 & 24.2 & 25.7 & 26.7 & 26.1 & 26.7 \\
SparQ        & 25.6 & 25.0 & 24.8 & 26.4 & 26.6 & 27.2 \\
StreamingLLM & 25.0 & 25.0 & 25.6 & 25.6 & 26.2 & 26.2 \\
H2O          & 24.8 & 24.8 & 25.2 & 25.8 & 26.2 & 27.0 \\
SnapKV       & 25.2 & 25.2 & 25.4 & 25.6 & 25.8 & 27.4 \\
\method      & 25.4 & 26.1 & 25.8 & 26.6 & 26.2 & 27.4 \\
\midrule
\multicolumn{7}{c}{\textit{Qwen3-8B} (Dense: 31.4)} \\
\midrule
Quest        & 5.6 & 9.1 & 15.1 & 17.5 & 22.5 & 24.7 \\
Multipole    & 29.4 & 30.0 & 28.4 & 29.8 & 28.2 & 28.8 \\
TokenSelect  & 28.7 & 29.1 & 29.3 & 29.5 & 29.5 & 29.5 \\
FreeKV       & 27.3 & 27.6 & 27.9 & 31.7 & 30.5 & 30.3 \\
SparQ        & 28.2 & 27.8 & 28.8 & 29.8 & 29.4 & 29.2 \\
StreamingLLM & 24.6 & 26.2 & 28.0 & 28.0 & 27.4 & 28.0 \\
H2O          & 26.2 & 26.6 & 27.8 & 28.0 & 28.2 & 29.2 \\
SnapKV       & 26.2 & 27.8 & 28.2 & 28.2 & 28.4 & 30.0 \\
\method      & 28.5 & 28.8 & 29.5 & 30.1 & 29.9 & 31.1 \\
\midrule
\multicolumn{7}{c}{\textit{Qwen3-14B} (Dense: 32.0)} \\
\midrule
Quest        & 9.3 & 18.1 & 21.5 & 24.7 & 29.0 & 30.6 \\
Multipole    & 30.4 & 29.2 & 29.2 & 28.0 & 28.8 & 28.8 \\
TokenSelect  & 29.6 & 30.4 & 30.8 & 30.4 & 30.6 & 30.8 \\
FreeKV       & 24.0 & 26.4 & 28.8 & 30.2 & 30.6 & 32.0 \\
SparQ        & 29.8 & 30.2 & 30.4 & 30.0 & 30.6 & 30.8 \\
StreamingLLM & 27.0 & 29.4 & 30.8 & 31.2 & 31.2 & 31.4 \\
H2O          & 28.6 & 30.4 & 31.2 & 31.4 & 31.8 & 31.2 \\
SnapKV       & 27.8 & 30.2 & 30.6 & 31.4 & 31.8 & 32.0 \\
\method      & 30.2 & 30.0 & 30.5 & 30.7 & 29.7 & 32.0 \\
\bottomrule
\end{tabular}
\end{table}

\subsection{RULER}

\begin{table}[!ht]
\caption{RULER accuracy for direct retrieval methods. MK-2, MQ, and VT denote Multikey-2, Multiquery, and variable tracking, respectively. Results use a matched 4,096-token retrieval budget. The average covers the three reported tasks only. Qwen3 tasks use 64k contexts; for Llama-3.1, the NIAH tasks use 64k contexts and variable tracking uses 32k.}
\label{tab:ruler_direct_retrieval}
\centering
\setlength{\tabcolsep}{3pt}
\begin{tabular}{@{}lrrrr@{\hspace{12pt}}rrrr@{}}
\toprule
\multicolumn{1}{c}{}
& \multicolumn{4}{c}{\textit{Qwen3-8B}}
& \multicolumn{4}{c}{\textit{Llama-3.1-8B-Instruct}} \\
\cmidrule(lr){2-5}\cmidrule(l){6-9}
\textbf{Method}
& \textbf{MK-2} & \textbf{MQ}
& \textbf{VT} & \textbf{Avg.}
& \textbf{MK-2} & \textbf{MQ}
& \textbf{VT} & \textbf{Avg.} \\
\midrule
CommunityKV & 78.0 & 92.5 & 22.0 & 64.2 & 95.0 & 97.8 & 96.8 & 96.5 \\
TokenSelect & 81.0 & 91.8 & 17.4 & 63.4 & 92.0 & 90.0 & 93.2 & 91.7 \\
FreeKV      & 64.0 & 93.5 & 46.4 & 68.0 & 91.0 & 97.0 & 98.4 & 95.5 \\
SparQ       & 69.0 & 88.5 & 26.8 & 61.4 & 47.0 & 85.5 & 92.0 & 74.8 \\
\bottomrule
\end{tabular}
\end{table}

\begin{table}[!ht]
\caption{RULER accuracy across the complete 13-task suite. Results use a matched 4,096-token active-token budget. Dense attention is included as a reference; reported values are unweighted averages over all 13 tasks.}
\label{tab:ruler_complete_suite}
\centering
\setlength{\tabcolsep}{5pt}
\begin{tabular}{@{}lrrrrr@{}}
\toprule
\textbf{Model} & \textbf{Dense} & \textbf{StreamingLLM}
& \textbf{H2O} & \textbf{SnapKV} & \textbf{CommunityKV} \\
\midrule
Qwen3-8B & 76 & 17 & 17 & 52 & 53 \\
Llama-3.1-8B-Instruct & 86 & 20 & 21 & 76 & 73 \\
\bottomrule
\end{tabular}
\end{table}

\end{document}